\documentclass[11pt,a4paper]{article}
\usepackage[hyperref]{eacl2021}
\usepackage{times}
\usepackage{latexsym}
\usepackage{wrapfig}
\usepackage{graphics}
\usepackage{graphicx}
\usepackage{multirow}
\usepackage{amsmath}
\usepackage{stfloats}
\usepackage{soul}
\usepackage{color,xcolor}
\sethlcolor{yellow}

\usepackage{microtype}

\aclfinalcopy % Uncomment this line for the final submission
\title{Modeling Coreference Relations in Visual Dialog}

\author{Mingxiao Li \\
  KU Leuven \\

  \texttt{mingxiao.li@kuleuven.be} \\\And
  Marie-Francine Moens \\
  KU Leuven  \\
  \texttt{sien.moens@cs.kuleuven.be} \\}

\date{}

\begin{document}
\maketitle
\begin{abstract}
Visual dialog is a vision-language task where an agent needs to answer a series of questions grounded in an image based on the understanding of the dialog history and the image. The occurrences of coreference relations in the dialog makes it a more challenging task than visual question-answering. Most previous works have focused on learning better multi-modal representations or on exploring different ways of fusing visual and language features, while the coreferences in the dialog are mainly ignored. In this paper, based on linguistic knowledge and discourse features of human dialog we propose two soft constraints that can improve the model's ability of resolving coreferences in dialog in an unsupervised way. Experimental results on the VisDial v1.0 dataset shows that our model, which integrates two novel and linguistically inspired soft constraints in a deep transformer neural architecture, obtains new state-of-the-art performance in terms of recall at 1 and other evaluation metrics compared to current existing models and this without pretraining on other vision-language datasets. Our qualitative results also demonstrate the effectiveness of the method that we propose. \footnote{Our code are released on: https://github.com/Mingxiao-Li/Modeling-Coreference-Relations-in-Visual-Dialog}
\end{abstract}

\section{Introduction}
Recently, with the unprecedented advances in computer vision and natural language processing, we have seen a considerable effort in developing artificial intelligence (AI) agents that can jointly understand visual and language information. Visual-language tasks, such as image captioning \cite{img_caption} and visual question-answering (VQA) \cite{vqa}, have achieved inspiring progress over the past few years. However, the applications of these agents in real-life are still quite limited, since they cannot handle the situation when continuous information exchange with a human is necessary, such as in  visual-language navigation \cite{vl_navigation} and visual dialog \cite{visualdialog}.
\begin{figure}
  %\vspace{-10pt}
  \begin{center}
    \includegraphics[width=0.5\textwidth]{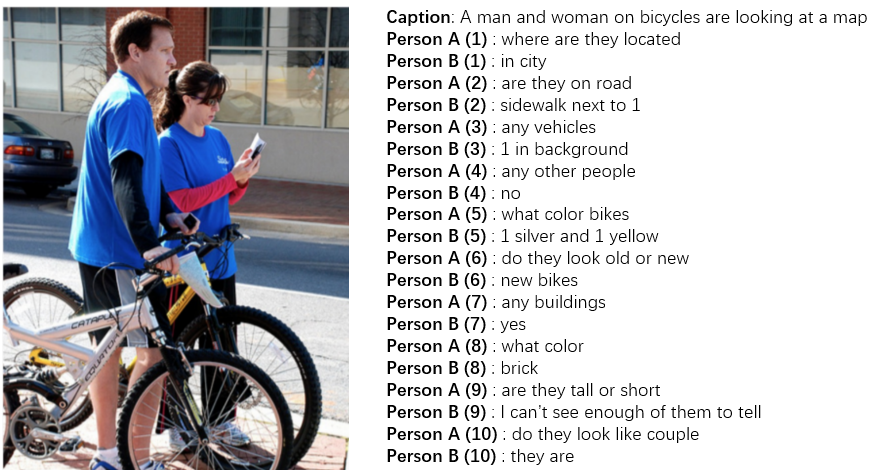}
  \end{center}
  %\vspace{-10pt}
  \caption{An example taking from the VisDial v1.0 dataset. The questioner (Person A) sees the caption and tries to understand the whole scene of the image by asking questions to the answerer (Person B) who can see the whole image.}
  %\vspace{-10pt}
\end{figure}
The visual dialog task can be seen as a generalization of VQA. Both tasks require the agent to answer a question expressed in natural language about a given image. A VQA agent needs to answer a single question, while a dialog agent has to answer a series of language questions based on its understanding of visual content and dialog history. Compared to VQA, the visual dialog task is more difficult because it demands the agent to resolve visual coreferences in the dialog. Considering the example in Figure 1.  when the agent encounters question 6 ``do they look old or new ?" and questions 9 ``are they tall or short ?", it has to infer that the pronoun ``they" in these two questions refers to different entities in the image or the dialog history. 

This paper studies how we can improve the results of a visual dialog task by better resolving the coreference relations in the dialog. In this work we restrict coreference resolution to pronouns. We use a multi-layer transformer encoder as our baseline model. Based on the assumption that the output contextual embedding of a pronoun and its antecedent should be close in the semantic space, we propose several soft constraints that can improve the model's capability of resolving coreferences in the dialog in an unsupervised way (i.e., without ground truth coreference annotations in the training data). Our first soft constraint is based on the linguistic knowledge that the antecedent of a pronoun can only be a noun or noun phrase. To integrate this constraint in the baseline model, we introduce a learnable part-of-speech (POS) tag embedding and a part-of-speech tag prediction loss. Inspired by the observation that in human dialog the referents of pronouns often occur in nearby dialog utterances, we propose a second soft constraint using a sinusoidal sentence position embedding, which aims to enhance local interactions between nearby sentences.     

Our contributions are as follows: First, as a baseline we adapt the multi-layer transformer encoder to the visual dialog task and obtain results comparable to the state of the art. Second, we propose two soft constraints to improve the model's ability of resolving coreference relations in an unsupervised way. We also perform an ablation study to demonstrate the effectiveness of the introduced soft constraints. Third, we conduct a qualitative analysis and show that the proposed model can resolve pronoun coreferents by making sure that in the neural architecture the pronoun mostly attends to its antecedent. 

\section{Related Work}
\textbf{Visual Dialog}. The Visual Dialog task is proposed by \newcite{visualdialog}, where a dialog agent has to answer questions grounded in an image based on its understanding of the dialog history and the image. Most of previous work focuses on using an attention mechanism to learn interactions between image, dialog history and question. \newcite{multi_step} use an attention network to conduct multi-step reasoning in order to answer a question. \newcite{recursive} propose a recursive attention network, which selects relevant information from the dialog history recursively. \newcite{dual} apply a multi-head attention mechanism \cite{attention} to learn mutimodal representations. \newcite{factor} fuse information from all entities including question, answer, dialog history, caption and image using a factor graph. \newcite{vd_pre} propose two-stage training. They first pretrain their transformer based two-stream attention network on other visual-language datasets, then finetune it on the visual dialog dataset. Other approaches consider different learning methodologies to model the visual dialog task, for example, \newcite{best} use adversarial learning and \newcite{rl} apply reinforcement learning.  

\textbf{Coreference Resolution.} Coreference resolution aims at detecting linguistic expressions referring to the same entities in the context of the discourse. The task has been dominated by machine learning approaches since the first learning based coreference resolution system was proposed by \newcite{connolly1997machine}. Before \newcite{end2endco} proposed the first end-to-end neural network based coreference resolution system, most of the learning-based systems have been built with hand engineered linguistic features. \newcite{easy_vectories} use surface linguistic features, such as mention type, the semantic head of a mention, etc., and their combinations to build a classifier to determine if two mentions refer to the same entity. \newcite{adapting} adopt integer linear programming (ILP) to introduce coreference constraints including centering theory constraints, direct speech constraints and definite noun phrase and exact match constraints in the inference step in order to adapt an existing coreference system trained on the newswire domain to short narrative stories without any retraining. Recently, \newcite{bertco} apply a BERT model to coreference resolution and achieve promising results on the OntoNotes corpus \cite{pradhan2012conll} and the GAP dataset \cite{gap}. Different from all coreference systems mentioned above, which rely on supervised learning and on a dataset annotated with coreference links, our work focuses on applying soft linguistic constraints to improve the model's ability of resolving coreferents in an implicit and unsupervised way. Similar to the work of \newcite{vision_co_entity} that operates on language and vision information, our model uses attention to jointly learn multi-modal representations.   

\begin{figure*}[h]
  %\vspace{-10pt}
  \centering
    \includegraphics[width=2.0\columnwidth]{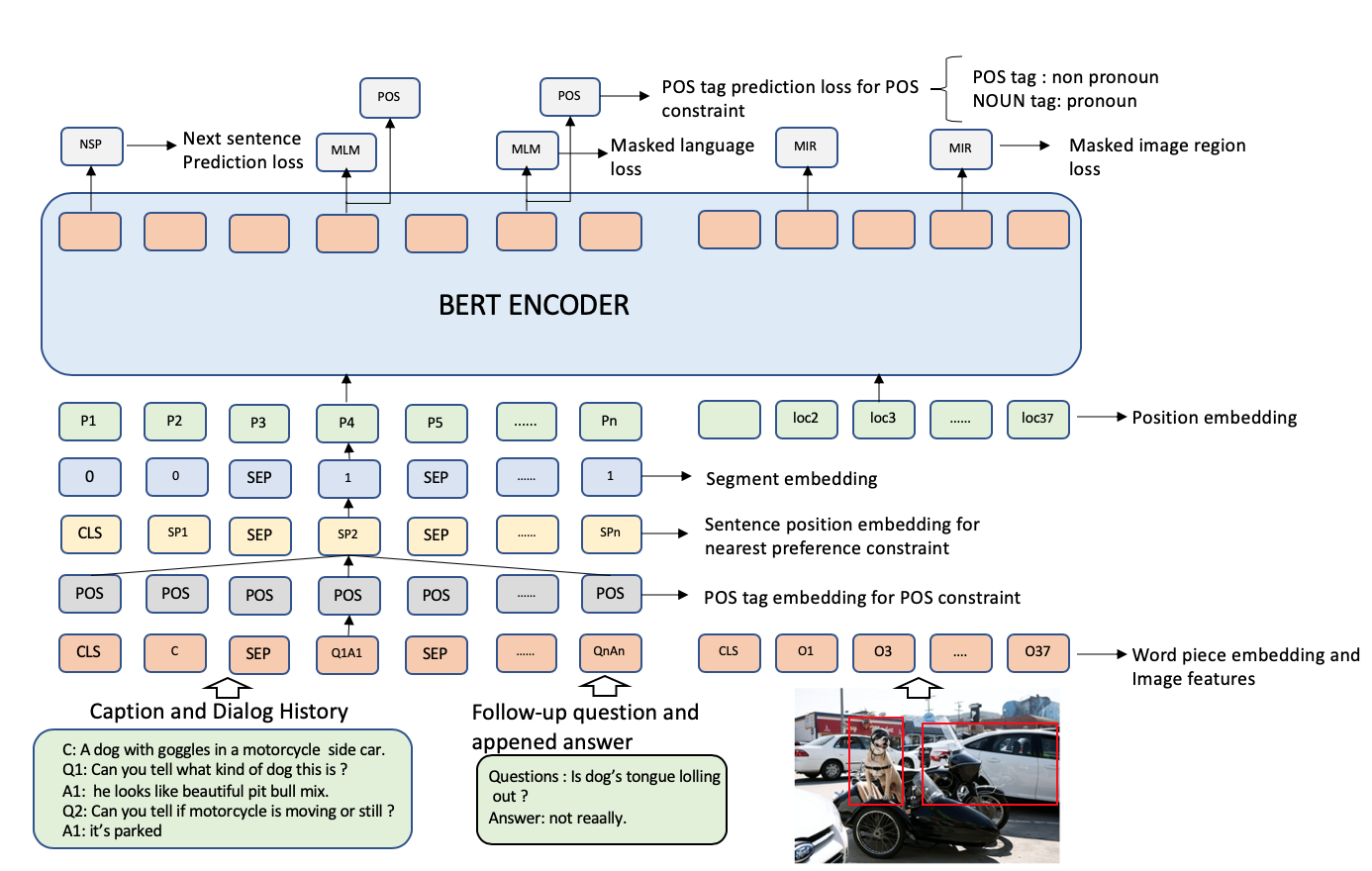}
  %\vspace{-10pt}
  \caption{The model architecture with the two soft constraints that we propose. The baseline model takes the image feature $I$, image caption $C$, dialog history $H_t$, follow-up question $Q_t$ and appended candidate answer $A_t$ as input and is trained by using a masked language model (MLM), masked image region (MIR) and next sentence prediction (NSP) losses. Two soft constraints are integrated into the model by adding POS tag embedding and sentence position embedding to the input language sequence embedding and by introducing a new POS tag prediction objective during training.}
  %\vspace{-10pt}
\end{figure*}

\section{Methodology}
\label{sec:methodology}
In this section, we formally describe the visual dialog task \cite{visualdialog} and the approaches we propose. In visual dialog, given an image $I$, the image caption $C$ and the dialog history until round $t-1$, $H=((Q_1,A_1),(Q_2,A_2),\cdots,(Q_{t-1},A_{t-1}))$, which is a sequence of question-answer pairs expressed in natural language and grounded in the image, a dialog agent is expected to correctly answer the question at round $t$ by choosing the answer from $100$ candidate answers $A_t = \{A_t^1,A_t^2,\cdots,A_t^{100}\}$. 

We first introduce the baseline visual dialog model in Section 3.1, followed by the detailed explanation of our proposed soft coreference constraints and how we integrate these into the baseline model in Section 3.2. The coreference constraints are based on linguistic knowledge, so consequently our method is an example of how to integrate linguistic knowledge into a neural transformer architecture. Figure 2 shows the architecture of our proposed model.

\subsection{Baseline Model}
\textbf{Transformer Encoder}. We use a multi-layer transformer \cite{attention} encoder as our baseline model. The computation within a single layer transformer encoder is presented in appendix A.1, and the details of the input and training objective functions are illustrated in below subsections. The main idea of applying the transformer architecture to the visual dialog task is to use the multi-head self-attention mechanism to implicitly learn the intra and inter interactions within the single modality and between the different modalities (in this case language and vision), respectively. 

\textbf{Linguistic Representation}. Following the monolingual BERT model \cite{bert} and the multi-modal BERT model \cite{vilbert,visualbert,vlbert}, we use the WordPiece \cite{wordpiece} tokenization tool to tokenize each input sequence into word pieces sequence. Then the sum of the word piece embedding, position embedding and segment embedding, where the segment embedding is used to differentiate questions from answers and to delimit boundaries of question-answer pairs, are taken as the language sequence input of the model. 

\textbf{Image Representation}. Following the multi-modal BERT model \cite{img_feature,vilbert,visualbert,vlbert}, we use Faster-RCNN \cite{faster_rcnn} with ResNet \cite{res_net} backbone to detect objects in the image and keep the top-36 detected objects and their corresponding bounding boxes. The representation of each detected object is obtained by applying a mean-pooled convolution on the region of that object. We also project a $5-d$ geometrical representation of each box, including the normalized top-left and bottom-right coordinates of the detected objects and the fraction of the area they cover, to the same dimensions as the image feature vector. In this way we obtain a vector with the same dimensions as the feature representation of the image. The final input image representation is the sum of its geometrical and feature representations. To avoid missing image information that is not captured by the top-36 bounding boxes, we also concatenate the mean-pooled feature vector of the whole image to the beginning of the image region sequence.  

\textbf{Multi-modal Input}. As the transformer encoder receives a sequence of tokens as input, to feed both image and language into the model, we simply concatenate the language sequence embedding and image region sequence representation to form a whole input sequence. Like in the BERT model, a special token [CLS] is added to the beginning of the input sequence to perform the next sentence prediction task. We also use another special token [SEP] to separate each question-answer pair and the two modalities. Our input sequence can be formulated as follows: $Input = \{[CLS],C,[SEP],Q_1A_1,[SEP],Q_2A_2,\cdots,Q_t$ $A_t,[SEP],O_0,O_1,O_2,\cdots,O_{36}\}$, where $C$ is the image caption. $Q_i$,$A_i$ are question and answer at round $i$, and $O_{0\sim36}$ denote the input image region features. 

\textbf{Multitasks Training Objectives}. To make the model learn a good alignment between different modalities, we utilize three losses: masked language model loss (MLM), masked image region loss (MIR), and next sentence prediction loss (NSP). Similar to the MLM in BERT, we randomly mask $15\%$ word pieces in the language sequence by replacing the word piece with a special token [MASK], while in MIR, we randomly set $15\%$ of the image region features to zero vectors. The model is trained to recover the masked words and predict the semantic category of the masked image regions:
\begin{align*}
    L_{MLM} &=  - E_{(I,w)\sim D}logP(w_m|w_{\backslash m},I) \tag{1}  \\
    L_{MIR} &= \sum_{i}^{k}KL(P_{m}||P_{g})  \tag{2}
\end{align*}
where $w_{\backslash m}$ and $I$ denote the word sequence excluding the masked words $w_m$, and image regions, respectively. KL represents the KL divergence loss. $P_m$ is the model output distribution and $P_g$ is the ground truth classification distribution. 

Recall that the visual dialog system aims to find the correct answer among the $100$ candidate answers. We realize this in a discriminative manner by using the next sentence prediction loss (NSP). We randomly select $1$ wrong answer from the candidate answers to generate negative samples, together with the ground-truth to form a balanced training dataset. During training, a candidate answer is appended to the dialog sequence, and the model is trained to predict whether or not the appended answer is the correct answer to the current question:
\begin{equation}
    L_{NSP} = -E_{(I,w)\sim D}logP(\Hat{y}|I,w) \tag{3}
\end{equation}
where $\hat{y}\in [0,1]$ is the output probability of the binary classifier at the last layer using the special [CLS] tag representation, which indicates the probability of the appended answer being correct. During inference, we rank the $100$ candidate answers using their NSP score, which is the $\hat{y}$ in the above equation. During training, the total loss is the sum of MLM, MIR and NSP losses:

\begin{equation}
    L_{total} = L_{MLM} + L_{MIR} + L_{NSP} \tag{4}
\end{equation}
\subsection{Soft Coreference Constraints}

As discussed before, the existence of pronouns in language makes the visual dialog a more challenging task than VQA. A naive way to reduce the difficulty would be to use a loss to guide the model to jointly learn to resolve the coreferences in the dialog and to generate an answer to the question. However, the lack of coreference annotations in the visual dialog dataset prevents from using this supervised learning approach. Although it is impossible to resolve coreferences directly in the model, we propose to use linguistic knowledge to improve the model's ability to implicitly resolve the coreferences in an unsupervised way. We do so by exploiting the attention mechanisms of the transformer architecture where attention weights act as soft constraints to guide the training of the model. The intuition behind it is the following. As the baseline model will output a contextual representation for each input token in the last layer, if a pronoun refers to a noun or noun phrase in the input sequence, the output contextual embedding of this pronoun and its antecedent should be close in the semantic space, which also means that the pronoun should attend most to its antecedent. 

\textbf{Part-Of-Speech Constraint}.
Our first proposed soft constraint is based on the linguistic knowledge that if the antecedent of a pronoun exists, it can only be a noun or noun phrase. We use the POS %Part-Of-Speech (POS) 
tag information and introduce a POS tag prediction loss 
%to implement this soft constraint 
to help pronouns to find nouns in the dialog. The Stanford CoreNLP POS tagger \cite{stanford} is used to obtain the POS tag of each word in the input dialog, and all sub-word splits from a word share the same POS tag. Similar to the word embedding, we use a learnable embedding for each POS tag, which is further summed with the word piece embedding, position embedding and segment embedding to form the input sequence embedding of the model. In POS prediction loss, similar to the MLM loss, we randomly mask $15\%$ of POS tag of input tokens.
Then, the model is trained to predict the ground truth POS tag\footnote{We use the POS tags used in Penn Treebank \cite{penntreebank}.} of non-pronoun masked words, while for masked pronouns we replace the ground truth PRP tag with NN tag forcing the model to learn the contextual pronoun embedding that is close to nouns in the semantic space. The POS prediction loss can be formulated as the equation below:  

\begin{align*}
&L_{POS} = -E_{(w,I)\sim D}(P_{non-pronoun}+P_{pronoun}) \\
&P_{non-pronoun}=-logP(POS(w)|w_{\backslash m},I) \\ 
&P_{pronoun} = logP(NN|w_{\backslash m},I)) \tag{5}
\end{align*}

where $w_{\backslash m}$ denote all unmasked words, and $D$ is the dataset. This soft constraint will make pronouns focus more on nouns instead of other words such as verb, adverb or adjective, etc. As it does not violate any linguistic rules, it will not introduce a bias to the language model.

\textbf{Nearest Preference Constraint}. Our second soft constraint is inspired by the observation that in human dialog a pronoun is more likely to refer to the noun that is close to it. For example, in the visual dialog shown in Figure 1, in round 9 the pronoun ``they" refers to the ``buildings", in round 6 ``they" refers to ``bikes". However, it is not always the case that pronouns refer to the noun closest in the previous utterances hence our soft constraint. 
%instead of other nouns farther in the dialog history such as the "bikes" in round 5, and the antecedent of the other "they" in round 6 is the "bikes" in the closest preceding round. 
In visual dialog, some pronouns refer to noun phrases that occur much earlier in the discourse - skipping a few rounds - as utterances are very short.
To integrate this preference into the model, we adapt the sinusoidal word position embedding proposed in \cite{attention} and introduce a sentence position embedding:
\begin{align*}
    PE_{pos,2i} &= \frac{1}{k}sin(pos/(M+10000^{\frac{2i}{d}})) \tag{6} \\
    PE_{pos,2i+1} &= \frac{1}{k}cos(pos/(M+10000^{\frac{2i}{d}}))  \tag{7}
\end{align*}
where $pos$ is the sentence position in the dialog, $d$ is the hidden state size and the $M$ is the maximum number of sentences, which is $21$ in this visual dialog task. $k$ is a scaling factor to control the local interactions brought by sentence position embedding and we use $k=100$. Compared to the original sinusoidal position embedding proposed in \cite{attention}, our sentence position embedding has one more scaling factor and one more element $M$ in the denominator, which aims at restricting the product of the sentence position embedding to be a monotonically decreasing function with respect to $|pos_1-pos_2|$.
\begin{align*}
    PE_{pos_1}\cdot PE_{pos_2}=\frac{1}{k^2}\sum_{i=0}^{\frac{d}{2}-1}cos(w_i \Delta pos)  \tag{8}
\end{align*}
where $w_i = 1/(M+\epsilon^{\frac{2i}{d}})$, and $\Delta pos$ denotes the distance between two positions. $\epsilon$ is a parameter, which makes the wavelength of the sinusoidal function in each dimension to form a geometrical progression.\footnote{Following  \newcite{attention}, we set $\epsilon = 10000$.} 
Since $\Delta pos \in [-M,M]$, $w_i \Delta pos \in [-1,1]$, it follows that $cos(w_i \Delta pos) == cos(w_i |\Delta pos|)$ which is monotonically decreasing in the region of $[0,1]$.The details of the derivation of equation 8 are presented in appendix A.2. The closer two sentences are, the larger of the product of their sentence position embedding, resulting in stronger local interactions between nearby sentences in the dialog. This soft constraint can be easily integrated into the model by adding the sentence position embedding to the input sequence embedding including word piece embedding, position embedding, segment embedding and POS tag embedding.  
\begin{table*}[h]
\begin{center}
\begin{tabular}{l|cccccc}
\hline  Model &  MRR $\uparrow$ & R@1 $\uparrow$ & R@5 $\uparrow$ & R@10 $\uparrow$ & Mean $\downarrow$  \\ \hline
LF \cite{visualdialog} & 55.42 & 40.95 & 72.45 & 82.83 & 5.95 \\
HRE \cite{visualdialog}& 54.16 &39.93 &70.45 &81.50 &6.41 \\
MN \cite{visualdialog} & 55.49 &40.98 & 72.30 & 83.30 &5.92 &\\
CorefNMN \cite{co_fvisual} & 61.50 & 47.55&78.10&88.80&4.51  \\
FGA \cite{factor} & 63.70 &49.58&80.98&88.55 &4.51 \\
RVA \cite{recursive} & 63.03 & 49.03 & 80.40 &89.83& 4.18 & \\
HACAN \cite{rl} &64.22& 50.88 & 80.63&89.45 & 4.20 \\
Synergistic \cite{guo2019image} & 62.20 & 47.90&80.43&89.95 &4.17 \\
DAN \cite{dual} &63.20& 49.63&79.75 & 89.35 & 4.30 & \\
Dual VD \cite{jiang2020dualvd} &63.23 & 49.25 &80.23&89.70&4.11 \\
CAG \cite{guo2020iterative} & 63.49 & 49.85 &80.63 &90.15&4.11 \\
\hline
Baseline Model &62.13 & 47.38 & 80.40 & 90.17 &4.09 \\
Model + POS Embedding Only & 64.20 &48.10 & 81.22 & 90.20 & 3.98 \\
Model + POS Loss only & 64.78 & 49.88 & 82.40 & 90.85 & 3.86 \\
Model + C1 & 65.44 & 51.20 & 83.38 & 92.03 &  3.64 \\
Model + C2 & 66.14 & 51.97 & 83.63 & 91.55 & 3.60 \\
Model + C1 + C2 &\bf 66.53 &\bf 52.63 &\bf 84.13 &\bf 92.50 &\bf 3.40 \\
\hline 
\end{tabular}
\end{center}
\caption{\label{font-table} Results of the visual dialog models on the VisDial v1.0 test set. C1 and C2 refer to the POS constraint and nearest preference constraint, respectively.($\uparrow$: the higher the better; $\downarrow$: the lower the better)}
\end{table*}

\section{Experiments}
\subsection{Dataset}
We use the real environment VisDial v1.0 dataset in this work. The VisDial v1.0 has $123$k, $2$k and $8$k dialogs for training, validation and test, respectively. Each dialog contains one image with its caption from the MS-COCO dataset \cite{microsoft}, and ten rounds of question-answer pairs, which were collected from the chatting log of one questioner and one answerer who both were discussing the image. For each question, except for the correct answer, the dataset also provides another $99$ candidate answers to form an answer pool from  which a model needs to select the relevant answer. Note that, although the VisDial v1.0 dataset also contains a small dense annotation set in which a relevance score is given to each candidate answer in the answer pool, we do not use this small dataset to finetune all our models, as we consider recall at 1 as main evaluation metric, and finetuning on this dense dataset could degrade the model's performance when measured by recall at 1 \cite{vd_pre}.

\subsection{Evaluation Metrics}
We evaluate our proposed models using three evaluation metrics: (1) mean reciprocal rank (MRR) \cite{mrr}; (2) recall @ k, that is, the existence of the ground truth response in the top-k ranked items of the response list generated by the model with $k$ = 1, 5, or 10; and (3) mean rank (Mean) of the ground truth response, that is, the average rank of the ground truth answer in the model's output ranked list (lower is better).

\subsection{Training Details}
Inspired by the open source code\footnote{https://github.com/vmurahari3/visdial-bert} of \newcite{vd_pre}, we have implemented our model using the PyTorch framework \cite{pytorch}. Our model has the same configuration as the BERT$_{\rm BASE}$ model, which contains 12 transformer layers and each layer has 12 attention heads with a hidden state size of $768$. We set the maximum input length to be $256$ including $37$ image features. All models were trained using the Adam \cite{adam} algorithm with a base learning rate of $5e^{-5}$. A linear learning rate decay schedule is employed to increase the leaning rate from $5e^{-6}$ to $5^{-5}$ over $30k$ iterations and decay to $5e^{-6}$ over $40k$ iterations. Together with negative samples, each image in the VisDial v1.0 dataset can generate $20$ samples for the NSP task. Since these samples are fairly correlated and following the work of \newcite{vd_pre}, we randomly sub-sample 8 out of these 20 during training. We use the validation set to decide when to stop training. The batch size is $32$ in all our experiments, and different from the work of \newcite{vd_pre} and \newcite{vilbert}, we do not pretrain our model on other vision-language datasets. 

\begin{table*}[h]
\begin{center}
\begin{tabular}{l|cccccc}
\hline
\hline
\multicolumn{6}{c}{Results tested on validation set with 2 coreferences (237 samples) } & \\
\hline  Model &  MRR $\uparrow$ & R@1 $\uparrow$ & R@5 $\uparrow$ & R@10 $\uparrow$ & Mean $\downarrow$ \\ 
\hline
Baseline Model & 65.06  &51.38  &82.42  &91.79  &3.77  \\
\hline
Model + C1 & 67.44 & 54.04 &\bf 84.54 & 92.58 & 3.38 \\
\hline 
Model + C2 & 67.19 & 54.03 & 84.30 & 92.08 & 3.55                                      \\
\hline 
Model + C1 + C2 & \bf 67.61 & \bf 54.45 & \bf 84.54 &\bf 92.87 &\bf 3.34 \\
\hline 
\hline
\multicolumn{6}{c}{Results tested on validation set with 4 coreferences. (138 samples)} \\
\hline  Model &  MRR $\uparrow$ & R@1 $\uparrow$ & R@5 $\uparrow$ & R@10 $\uparrow$ & Mean $\downarrow$ \\ 
\hline
Baseline Model & 62.94  &48.93  &80.78  &90.28  &3.99  \\
\hline
Model + C1 & 65.77 & 51.99 & 83.49 &\bf 93.00 & 3.48 \\
\hline 
Model + C2 &  65.94 & 52.07 & 83.36 & 92.00 & 3.52  \\
\hline 
Model + C1 + C2 &\bf 66.53 &\bf 52.78 &\bf 84.21 & 92.07 & \bf 3.46 \\ 
\hline 
\hline
\multicolumn{6}{c}{Results tested on validation set with 6 coreferences. (58 samples)} \\
\hline  Model &  MRR $\uparrow$ & R@1 $\uparrow$ & R@5 $\uparrow$ & R@10 $\uparrow$ & Mean $\downarrow$ \\ 
\hline
Baseline Model & 60.02  &43.99  &80.80  &90.40  &4.06  \\
\hline
Model + C1 & 62.44 & 46.40 &\bf 83.86 & 92.66 & 3.61 \\
\hline 
Model + C2 & 63.53 &47.09 & 83.33 & 92.28 & 3.59  \\
\hline 
Model + C1 + C2 & \bf 63.66 & \bf 48.14 & 83.33 & \bf 92.67 &\bf 3.42 \\ 
\hline 
\hline 
\multicolumn{6}{c}{Results tested on validation set with coreferences. (1227 samples)} \\
\hline  Model &  MRR $\uparrow$ & R@1 $\uparrow$ & R@5 $\uparrow$ & R@10 $\uparrow$ & Mean $\downarrow$ \\ 
\hline
Baseline Model & 64.62  &50.73  &82.40  &91.42  &3.73  \\
\hline
Model + C1 & \bf 66.29 & 52.41 & 84.17 & 92.08 & 3.62 \\
\hline 
Model + C2 & \bf 66.29 & 52.54  & 83.46 & 91.25 & 3.59  \\
\hline 
Model + C1 + C2 & 65.97 & \bf 53.27 & \bf 83.47 & \bf 92.41 &\bf 3.48 \\ 
\hline 
\hline 
\end{tabular}
\end{center}
\caption{\label{font-table} Results of the visual dialog models obtained on three small datasets each with a different number of pronoun coreferences that were collected from the VisDial v1.0 validation set. C1 and C2 refer to the POS constraint and nearest preference constraint, respectively.}
\end{table*}

\section{Results }
\subsection{Quantitative Results}

We compare the results of our model with the results of the following previously published models obtained on the VisDial v1.0 dataset: LF \cite{visualdialog}, HRE \cite{visualdialog}, MN \cite{visualdialog}, CorefNMN \cite{co_fvisual}, FGA \cite{factor}, RVA \cite{recursive}, HACAN \cite{rl}, Synergistic \cite{guo2019image}, DAN \cite{dual}, Dual VD \cite{jiang2020dualvd}, and CAG \cite{guo2020iterative}. To make a fair and transparent comparison, we do not compare our models with models which were pretrained on other vision-language datasets before finetuning them on the VisDial v1.0 dataset, all the more because the vision-language datasets used in the pretraining overlap with the testset of Visdial v1.0. Also for those models, such as FGA, for which the authors also provide results of ensemble models, we only consider the results of their single model.

\textbf{Results on VisDial v1.0 testset}. As presented in Table 1, our best model (baseline model with both soft constraints) significantly outperforms all the previous published models and reach new state-of-the-art performance on MRR, R@1, R@5 and R@10. Specifically, compared to the best performance of previous models, our best model improves around $2.3\%$ on MRR, $1.78\%$ on R@1, $2.13\%$ on R@5 and $2.35\%$ on R@10. The mean rank of our proposed model is also better than all previous models, although the difference is relative small around $0.71$. We also tested all our models on the VisDial v1.0 development set, and the results are presented in appendix A.3.

\textbf{Ablation study}. To further study the effectiveness of the two soft constraints, we perform an ablation study on the VisDial v1.0 dataset with four different models: (1) Baseline model; (2) Model with the POS soft constraint (Model + C1); (3) Model with the nearest preference constraint (Model + C2); (4) Model with both the POS and nearest preference constraints (Model + C1 + C2). Moreover, We conduct an ablation study for the two aspects (POS embedding and POS prediction loss) in POS constraint The results are presented in Table 1. The baseline model obtains the following results in terms of MRR ($62.13\%$), R@1 ($47.38\%$), R@5 ($80.40\%$) and R@10 ($90.17\%$). Models with only POS embedding and only POS prediction loss have better performance than the baseline model. Further combining both leads to our first POS constraint, which improve the performance across all evaluation metrics ($3.31\%$ for MRR, $3.82\%$ for R@1, $2.98\%$ for R@5, $1.86\%$ for R@10 and $0.55$ for MRR). Similarly, only considering the nearest preference constraint leads to better performance on all evaluation metrics. The last row of Table 1 illustrates that the proposed soft constraints jointly lead to better results. We also study the changes of attention distribution of the model with and without our proposed constraints. Figure 3 shows that the nearest constraint can enhance the local connection in dialog, and the POS constraint is able to make pronouns focus more on nouns. These results indicate the effectiveness of adding linguistic constraints to a neural network. Integrating linguistic knowledge in a transformer neural architecture effectively improves the model's performance in the visual dialog task.

\begin{figure}[h]
    \centering 
    \includegraphics[width=1.0\columnwidth]{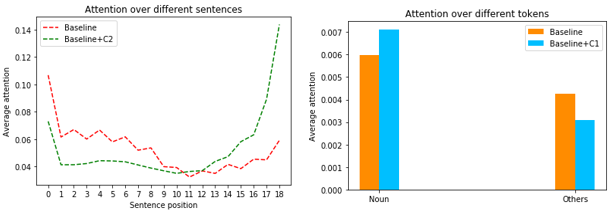}
    \caption{Left: Attention distribution of question over dialog history (Baseline Model+C2). Right: Attention distribution of pronoun over nouns and other words. (Baseline Model+C1}
    \label{fig:my_label}
\end{figure}

\begin{figure}[h]
    \centering 
    \includegraphics[width=1.0\columnwidth]{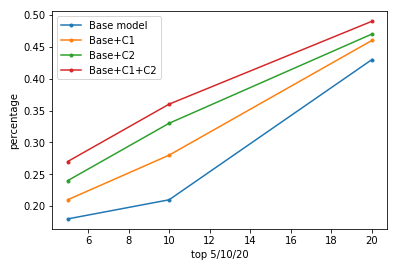}
    \caption{The percentage of the correct antecedent is within top 5/10/20 pronoun's attention distribution.}
    \label{fig:my_label}
\end{figure}

\begin{figure*}[h]
    \centering 
    \includegraphics[width=1.8\columnwidth]{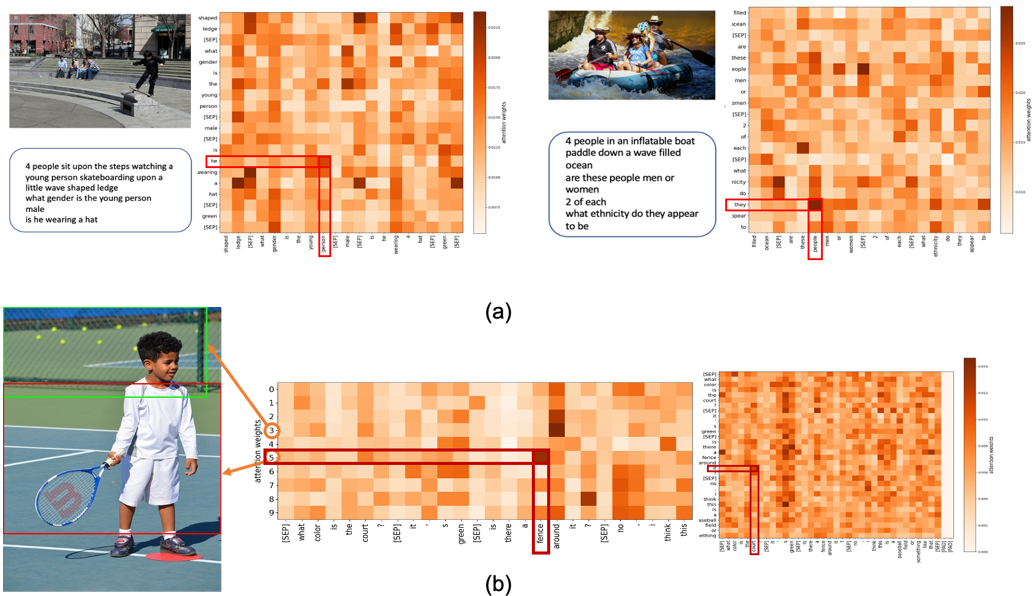}
    \caption{Attention score map of our best model (Model C1+C2). (a) Two correct cases: The attention map shows that the pronoun correctly attends to its antecedent. (b) An incorrect case: Although the coreference is correctly resolved, the word "fence" does not attend to the correct region in the image, which results in selecting the wrong answer.}
    \label{fig:my_label}
\end{figure*}

\subsection{Coreference Analysis}
As we do not have the access to the ground truth of the VisDial v1.0 test set, to further analyze the effectiveness of the proposed models, we create three small datasets, which each consists of samples with $2$, $4$, and $6$ coreferences, respectively, from the Visdial v1.0 validation set. As there is no coreference annotation in the Visdial v1.0 dataset, we assume that each third-person pronoun (he, she, they, him, her, them) and possessive pronouns (its, his, her, their) has one coreference in the dialog.\footnote{Note that the dialog is about the objects in an image.} To create these subsets, we do not take ``it" into consideration, as many of the occurrences of ``it" \footnote{``It" has many other functions apart from being a pronoun, such as ``empty" subject or object, not referring to anything in particular, to introduce or anticipate the subject or object of a sentence, use in cleft or in passive voice sentences, etc.} do not have a corefering expression, for example, ``is it daytime ?". We test our models and the baseline model using these four small datasets, and the results are illustrated in Table 2. Comparing the results in Table 2 with that in Table 1, in some cases the performance in Table 2 is better, which means that the difficulty of these sampled data do not higher than that of the test set. One clear observation is that in almost all cases the model with two soft constraints has the best performance in terms of MRR, R@1, R@5, R@10 and Mean in all three datasets. Another observation is that integrating either the POS constraint or the nearest preference constraint improves the performance across all evaluation metrics in all four datasets, which again shows the effectiveness of our proposed linguistically inspired soft constraints. We can also see the trend that the models' performance is worse when the dialog has more coreferences, which is reasonable and consistent with our previous assumption that the existence of coreferences makes this task more difficult. To further study the effectiveness of our proposed constraints, we manually annotated coreferences in a subset of the validation set and tested all our ablation models on this dataset. The results are presented in Figure 4, which shows that both the proposed constrains can improve the model's ability of resolving coreferences.

\subsection{Qualitative Analysis}
We visualize the attention scores of the model to investigate whether or not our best model can resolve pronoun coreferences in the dialog. We expect that the pronoun will attend the most to its antecedent, if the model correctly resolves the pronoun referent. Figure 5(a) presents the results of two samples taken from the VisDial v1.0 test set. Note that we only show the attention scores in the window of size 20 around the pronoun and its possible antecedent. The attention score maps illustrate that in these examples the attention weights between pronoun and its antecedent are significantly larger than those between the pronoun and other words, which indicates that the model can implicitly resolve the coreferences correctly.

\textbf{Error Analysis} Figure 5(b) presents a negative example. When facing the question ``Is there a fence around it ?", the model answers ``no, i think..." instead of the correct answer ``yes". As shown in the attention map within language sequences, the pronoun ``it" does attend to its antecedent ``court", implying that the model successfully resolves the corefering noun. However, looking at the cross-modal attention, the word ``fence" incorrectly refers to region 5 (court) in the image, which results in selecting a wrong answer. This negative example indicates that enhancing the model's ability of correct visual grounding is a meaningful future work.

\section{Conclusion}
In this paper, we have built a multi-layer transformer model for the visual dialog task. Based on linguistic knowledge and human dialog discourse patterns, we have proposed two soft constraints that effectively improve the model's performance by enhancing its ability of implicitly resolving pronoun coreferences. We have used the VisDial v1.0 dataset to evaluate our model. Our model obtains new state-of-the-art performance in correctly answering the dialog questions when compared to existing models without pretraining on other vision-language datasets. Our coreference and qualitative analysis further supports the proposed soft constraints.  

\section{Acknowledgement}
This research received funding from the Flemish Government (AI Research Program).

\bibliography{anthology,eacl2021}
\bibliographystyle{acl_natbib}

\clearpage
\setcounter{table}{0}

\appendix
\section{Appendix}

\appendix
\section{1}
This Appendix shows the computation within a transformer encoder layer. Given an input sequence $H_0=[e_0,e_1,\cdots,e_n]$, the transformer encodes it into different levels of contextual representations using a multi-head self-attention mechanism:
\begin{align*}
&Q = H_{t-1}W_{t}^Q, K = H_{t-1}W_{t}^K, V=H_{t-1}W_{t}^V \tag{1} \\
&AttentionHead_i(Q,K,V) = Softmax(\frac{QK^T}{\sqrt{d}})V  \tag{2}\\
&MultiHeadSelfAttention(H^t) = \\
&Concat(head_1,\cdots,head_k)W^O \tag{3}
\end{align*}

where $t$ is the layer index, $W_{t}^Q,W_{t}^K,W_{t}^V,W_{t}^O$ are learnable projection matrices, which map the hidden state $h_t$ to the query $q$ vector, key $k$ vector, value $v$ vector and output vector, and $H_{t}$ denote the learned contextual representations at layer $t$. $d$ is the size of hidden state. 

\appendix
\section{2}
The below formulations express the derivation of getting equation 8. Here, we use $p$ to denote the $pos$ in equation 11.
\begin{align*}
PE_{1}\cdot PE_{2}&= \frac{1}{k^2}\sum_{i=0}^{\frac{d}{2}-1}sin(w_i p_1)\cdot sin(w_i p_2) \\   &+ cos(w_i p_1)\cdot cos(w_i p_2) \\
&=\frac{1}{k^2}\sum_{i=0}^{\frac{d}{2}-1}cos(w_i(p_1-p_2)) \\
&=\frac{1}{k^2}\sum_{i=0}^{\frac{d}{2}-1}cos(w_i \Delta p) 
\end{align*}

\appendix
\section{3}
\begin{table*}[hp]
\begin{center}
\begin{tabular}{l|cccccc}
\hline  Model & \ MRR $\uparrow$& R@1 $\uparrow$ & R@5 $\uparrow$ & R@10 $\uparrow$ & 
Mean $\downarrow$ \\ \hline
Baseline Model &64.80 & 50.56 & 82.64 & 91.02 &3.85 \\
Baseline Model + C1 & 68.59 & 55.37 & 84.38 & 92.29 &  3.28 \\
Baseline Model + C2 & 68.32 & 55.25 & 84.49 & 92.25 & 3.32 \\
Baseline Model + C1 + C2 &\bf 69.49 &\bf 56.46 &\bf 85.33 &\bf 93.37 &\bf 3.19 \\
\hline 
\end{tabular}
\end{center}
\caption{\label{font-table} Results of the visual dialog models on the VisDial v1.0 development set. C1 and C2 denote our proposed soft constraints: POS constraint and nearest preference constraint, respectively.}
\end{table*}
Table 1 in this section shows the results of the four models on the VisDial v1.0 development set. Similar to the results obtained by testing on the test set, the model with both constraints (Model + C1 + C2) has the best performance across all evaluation metrics, and adding any one of the proposed soft constraint improves the performance of the baseline model, which indicates the effectiveness of our approach also during training.  

\label{sec:appendix}

%\section{Supplemental Material}

\end{document}